\documentclass[a4paper,10pt,twocolumn]{article}
\usepackage[utf8]{inputenc}
\usepackage[german,english]{babel}
\usepackage[T1]{fontenc}

\usepackage{cite}
\usepackage{amsmath,amssymb,amsfonts}
\usepackage{graphicx}
\usepackage{textcomp}
\usepackage{xcolor}
\def\BibTeX{{\rm B\kern-.05em{\sc i\kern-.025em b}\kern-.08em
    T\kern-.1667em\lower.7ex\hbox{E}\kern-.125emX}}
    
\usepackage{algorithm}
\usepackage{algpseudocode}
\usepackage{array}
\usepackage{amsthm}
\usepackage{graphicx}

\usepackage{color}
\definecolor{ourbrown}{RGB}{155,100,15}
\definecolor{ourpurple}{RGB}{145,0,140}
\definecolor{darkgreen}{RGB}{0,170,0}
\definecolor{darkorange}{RGB}{225,100,0}

\usepackage{xspace}

\newcommand{\nat}{\ensuremath{\mathbf{N} }}

\newcommand{\real}{\ensuremath{\mathbf{R} }}

\graphicspath{{figures/}}

\newtheorem{assumption}{Assumption}
\begin{document}

\title{Dimensionality Reduction and Anomaly Detection for \\
CPPS Data using Autoencoder}

\author{
Benedikt Eiteneuer$^1$, benedikt.eiteneuer@hs-owl.de\\
Nemanja Hranisavljevic$^2$, nemanja.hranisavljevic@iosb-ina.fraunhofer.de\\
Oliver Niggemann$^2$, oliver.niggemann@iosb-ina.fraunhofer.de\\
\ \\
$^1$Institute Industrial IT, Lemgo, Germany\\
$^2$Fraunhofer IOSB-INA, Lemgo, Germany\\
}

\date{February 2019 \\ \ \\
Copyright IEEE 2019, \\
International Conference on Industrial Technology (ICIT) 2019, \\
DOI: 10.1109/ICIT.2019.8755116}

\maketitle


\begin{abstract}
Unsupervised anomaly detection (AD) is a major topic in the field of Cyber-Physical Production Systems (CPPSs). A closely related concern is dimensionality reduction (DR) which is: 1) often used as a preprocessing step in an AD solution, 2) a sort of AD, if a measure of observation conformity to the learned data manifold is provided.

We argue that the two aspects can be complementary in a CPPS anomaly detection solution. In this work, we focus on the nonlinear autoencoder (AE) as a DR/AD approach. 
The contribution of this work is: 1) we examine the suitability of AE reconstruction error as an AD decision criterion in CPPS data. 2) we analyze its relation to a potential second-phase AD approach in the AE latent space 3) we evaluate the performance of the approach on three real-world datasets. Moreover, the approach outperforms state-of-the-art techniques, alongside a relatively simple and straightforward application.
\end{abstract}

\section{Introduction}
\label{intr}

Modern production systems are perceived as holistic and complex systems of many mechanical and computational elements as well as other aspects. This conception is accented in research agendas such as Cyber-Physical Production Systems (CPPSs) and "Industrie 4.0" (I4.0). \cite{Monostori:2014, Rajkumar:2010:CSN:1837274.1837461}

Computational elements of a CPPS can access a large number of variables, which typically describe system behavior, system environment, and product features. Contained information can be utilized to bring diverse benefits to the system such as robustness or resource efficiency. For this reason, data analysis increasingly gets attention in this field. 


One of the major expectations towards CPPSs is self-diagnosis, whose key task is the detection of anomalous behavior \cite{NL15}.
The authors assert that the future of anomaly detection for CPPSs is in data-driven model-based approaches. 
Such approaches create behavioral models from "normal" data and perform anomaly detection by comparing new data against the model. 
Herein lies the importance of machine learning (ML) in this domain---it can be used to learn such models. 

Typically, data anomalies can indicate a fault in the system, suboptimal behavior (e.g. suboptimal energy consumption) or wear of some components. Detection of such behavior can improve the reliability, safety, and efficiency of the system.

Today, the number of signals observed from some production system can often reach several hundred or thousands \cite{Jakubek:2002, NL15}.
Analysis of the system behavior, either by a human or a computer algorithm can be difficult when working with such high-dimensional data. 
Traditional AD (and other ML) approaches often face difficulties in handling high-dimensional datasets due to the curse of dimensionality \cite{Donoho:2000, Hinneburg:2000}. In order to overcome the dimensionality challenge and enable the use of the approaches that are suitable for fewer dimensions, dimensionality reduction can be performed. 

In this work, we analyze a concept based on neural-network autoencoder as a solution to the addressed challenges. Main hypotheses rely on the autoencoder dual nature: it performs dimensionality reduction and provides anomaly detection decision criterion (reconstruction error).

The rest of this paper is structured as follows:
Section \ref{sec:problem} declares the problem and gives related work. In Section \ref{sec:approach}, the proposed dimensionality reduction, and anomaly detection concept are described. Evaluation of the approach using real-world datasets is in Section \ref{sec:evaluation} while the conclusions and the future work are given in Section \ref{sec:conclusion}.





\section{Problem statement and related work}
\label{sec:problem}

\subsection{Dimensionality reduction}
Dimensionality reduction (DR) is the transformation of data observations into a meaningful representation of lower dimensionality \cite{van2009dimensionality}. The question arises: \emph{What is a meaningful representation of a reduced dimensionality?}

One way to define DR problem is the following: For a given $p, m \in \nat, p<m$, the DR is to find an encoding function $f_{enc}: \real^m \mapsto \real^p$ and a decoding function $f_{dec}: \real^p \mapsto \real^m$, which minimize the overall error between original observations and their reconstructions. Here, $m$ is the dimensionality of the input observations and $p$ is the reduced dimensionality.
Hopefully, the latent representation preserves meaningful features of the original observation, as a result of the low reconstruction error, commonly defined as:
\begin{equation}
\label{eq:l2loss}
MSE = \frac{1}{n}\sum_{i=1}^{n} |x_i - y_i|^2 
\end{equation}
where $|\cdot|$ denotes the standard $L^2$ norm and $n \in \nat$ is the number of data observations.

One of the simplest models is obtained when $f_{enc}(x) = Bx$, $B \in \real^{p \times m}$ and $f_{dec}(z) = Az$, $A \in \real^{m \times p}$ i.e. encoding and decoding function are matrix multiplications. The error function of this linear autoencoder \cite{baldi1989neural} becomes $MSE = \frac{1}{n}\sum_{i=1}^n|x_i-ABx_i|^2$.
It is apparent, that there is no unique solution for $A$ and $B$ as $AB = (AC)(C^{-1}B)$ for any invertible $C \in \real^{p \times p}$. 
The linear autoencoder can be represented as a neural network with a $p$-dimensional hidden layer and an $m$-dimensional output layer.

Principal component analysis (PCA) is a linear method to transform (encode) data into a new representation of linearly uncorrelated variables (principal components), defined by the eigenvectors of the data covariance matrix. 
Following the previously defined autoencoder framework, PCA is obtained when $f_{enc}(x) = Wx$ and $f_{dec}(z) = W^Tz$, $W \in \real^{p \times m}$. The weight matrix $W$ is composed of $p$ eigenvectors of the data covariance matrix corresponding to the $p$ largest eigenvalues. Such $W$ minimizes Equation \ref{eq:l2loss} \cite{baldi1989neural}. Apparently, PCA is similar to the linear autoencoder. \cite{baldi1989neural} proves that the unique global and local minimum in terms of linear AE parameters $A$ and $B$ is obtained when $AB$ is the orthogonal projection to the space spanned by the first $p$ principal components of the data.
 
 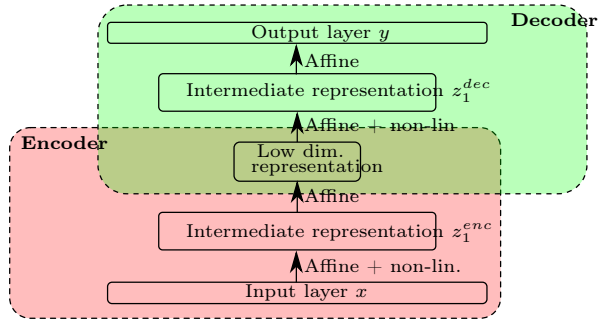
\begin{figure}
\setlength{\unitlength}{1\columnwidth}
\begin{picture}(0.95,0.55)
\put(0.00,0.00){\includegraphics[width=0.98\columnwidth]{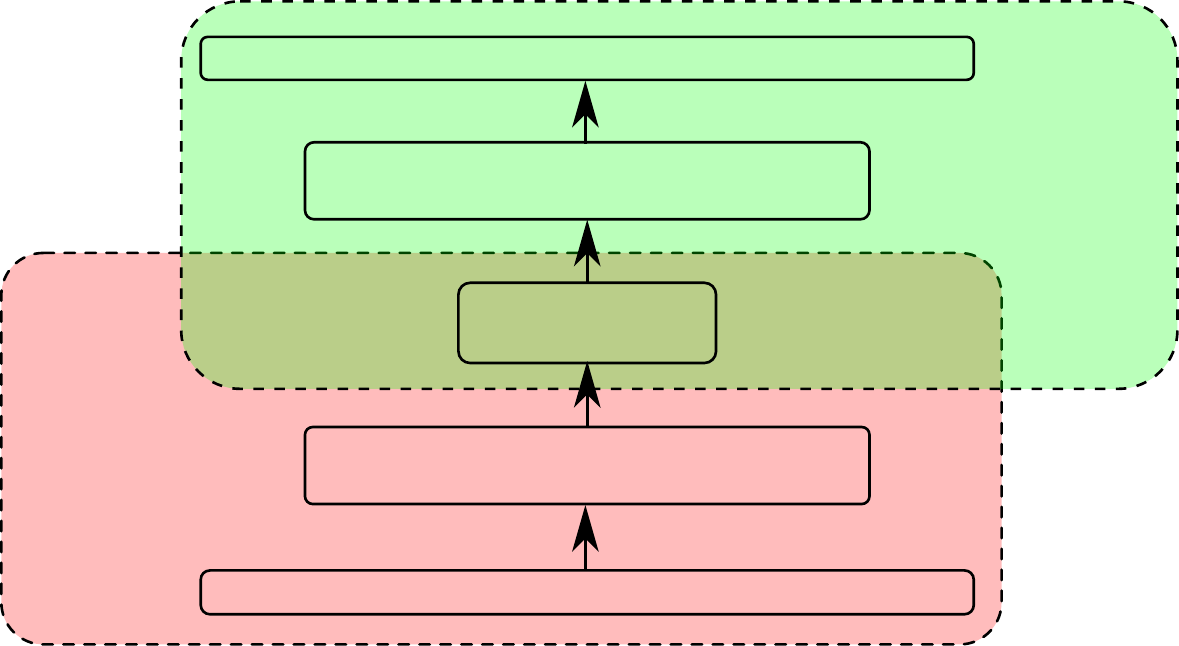}}

\put(0.85,0.50){\scriptsize{\textbf{Decoder}}}
\put(0.02,0.29){\scriptsize{\textbf{Encoder}}}

\put(0.41,0.48){\scriptsize{Output layer $y$}}
\put(0.30,0.38){\scriptsize{Intermediate representation $z^{dec}_1$}}
\put(0.42,0.276){\scriptsize{Low dim.}}
\put(0.41,0.250){\scriptsize{representation}}
\put(0.30,0.14){\scriptsize{Intermediate representation $z^{enc}_1$}}
\put(0.4,0.037){\scriptsize{Input layer $x$}}

\put(0.50,0.43){\scriptsize{Affine}}
\put(0.50,0.32){\scriptsize{Affine + non-lin}}
\put(0.50,0.20){\scriptsize{Affine}}
\put(0.50,0.08){\scriptsize{Affine + non-lin.}}
\end{picture}
\caption{Schematic description of a 4-layer autoencoder. Arrows represent the affine transformation and the model parameter. Boxes represent the different layer data representations.}\label{fig:ae_skizze}
\end{figure}


An autoencoder neural network or autoencoder \cite{hinton2006reducing} is a special type of deep feed-forward neural network, typically used for representation learning and dimensionality reduction. It utilizes the expressive power of a neural net by extending the previously described linear autoencoder with multiple layers of nonlinearities and affine transformations (see Figure \ref{fig:ae_skizze}). 
Nonlinear activation functions allow for non-linear feature extraction and modeling of arbitrarily complex functions. If the true underlying latent space is only accessible by such transformations, nonlinear techniques have to be used.
However, training an autoencoder with nonlinear activation functions is a non-convex problem which renders optimization non-trivial (back-propagation is commonly used). 
Development of deep learning, and particularly advances in unsupervised learning algorithms and network architectures, make autoencoder a convenient nonlinear DR technique \cite{hinton2006reducing,vincent2010stacked}.

\subsection{Anomaly detection}
Anomaly detection (AD) is a process of detecting observations (patterns) which do not conform to the expected (normal) behavior of the system \cite{Chandola:2009}. It stands to question: \textit{What is the normal behavior?}

Different techniques define normal behavior in different ways which makes them suitable for different problems. E.g. nearest-neighbor-based techniques \cite{Chandola:2009} assume that the normal-behavior observations occur in dense neighborhoods, while anomalies do not have close (normal) neighbors. On the other hand, DR approaches consider observations normal when they lay close to the learned low dimensional manifold.

\begin{figure}[tb]
  \begin{scriptsize}
    \centering  
    \def\svgwidth{230pt}
\begingroup%
  \makeatletter%
  \providecommand\color[2][]{%
    \errmessage{(Inkscape) Color is used for the text in Inkscape, but the package 'color.sty' is not loaded}%
    \renewcommand\color[2][]{}%
  }%
  \providecommand\transparent[1]{%
    \errmessage{(Inkscape) Transparency is used (non-zero) for the text in Inkscape, but the package 'transparent.sty' is not loaded}%
    \renewcommand\transparent[1]{}%
  }%
  \providecommand\rotatebox[2]{#2}%
  \newcommand*\fsize{\dimexpr\f@size pt\relax}%
  \newcommand*\lineheight[1]{\fontsize{\fsize}{#1\fsize}\selectfont}%
  \ifx\svgwidth\undefined%
    \setlength{\unitlength}{167.51631433bp}%
    \ifx\svgscale\undefined%
      \relax%
    \else%
      \setlength{\unitlength}{\unitlength * \real{\svgscale}}%
    \fi%
  \else%
    \setlength{\unitlength}{\svgwidth}%
  \fi%
  \global\let\svgwidth\undefined%
  \global\let\svgscale\undefined%
  \makeatother%
  \begin{picture}(1,0.48196799)%
    \lineheight{1}%
    \setlength\tabcolsep{0pt}%
    \put(0,0){\includegraphics[width=\unitlength,page=1]{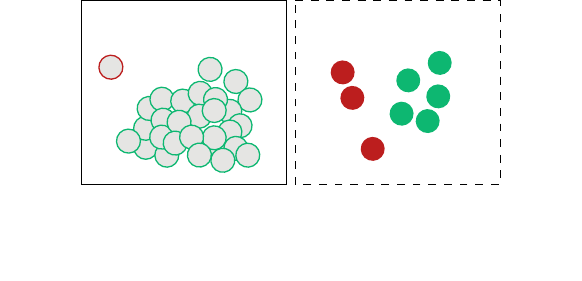}}%
    \put(0.18688606,0.44159147){\makebox(0,0)[lt]{\lineheight{1.25}\smash{\begin{tabular}[t]{l}Unlabeled training \end{tabular}}}}%
    \put(0.54282154,0.43263712){\makebox(0,0)[lt]{\lineheight{1.25}\smash{\begin{tabular}[t]{l}Small labeled dataset\end{tabular}}}}%
    \put(0.10865458,0.07621139){\makebox(0,0)[lt]{\lineheight{1.25}\smash{\begin{tabular}[t]{l}Test data\end{tabular}}}}%
    \put(0.74227122,0.07621139){\makebox(0,0)[lt]{\lineheight{1.25}\smash{\begin{tabular}[t]{l}Result\end{tabular}}}}%
    \put(0.26366784,0.41211599){\makebox(0,0)[lt]{\lineheight{1.25}\smash{\begin{tabular}[t]{l}dataset\end{tabular}}}}%
    \put(0,0){\includegraphics[width=\unitlength,page=2]{anomalies.pdf}}%
    \put(0.40810153,0.03259653){\makebox(0,0)[lt]{\lineheight{1.25}\smash{\begin{tabular}[t]{l}AD approach\end{tabular}}}}%
    \put(0,0){\includegraphics[width=\unitlength,page=3]{anomalies.pdf}}%
  \end{picture}%
\endgroup%

    \vglue 0ex plus 0.5ex minus 3.5ex
    {\caption{ Typical anomaly detection mode in CPPS. The approach uses a training set which is close to anomaly-free. Sometimes a small labeled dataset is available which can be used for choosing an optimal threshold parameter.}\label{anomalies}}
    \end{scriptsize}
  \end{figure}

Considering the typical characteristics of CPPSs, we are focused on a semi-supervised anomaly detection problem (categorization from \cite{Chandola:2009}). In this mode, the expected behavior is learned from a set of (mostly) normal observations which is common for AD problems in CPPSs, since the available data usually represent normal functioning of the production system. Then, a measure of abnormality provided by the solution is used to set a threshold for the anomaly detection (see Figure \ref{anomalies}). If a small labeled subset of data exists, it could be used to set a more motivated threshold. However, this should be considered with much care, because anomalies are intrinsically diverse in nature. This means, recorded anomalies might not represent well the possible anomalies one could encounter.

%


A traditional CPPS anomaly detection method is based on PCA, where reconstruction loss (Equation \ref{eq:l2loss}) of a test point serves as an anomaly score. To classify it as an anomaly one checks whether the score exceeds some predefined threshold. 
It could happen that anomalous points are very close to the learned manifold similar to normal data, but they still differ within the reduced space. Such points can only be detected by applying a \textit{second phase} anomaly detection on the reduced data (e.g. neighborhood-based, clustering or statistical anomaly detection methods). 

Two-phase approaches are common in the CPPS field. In the domain of automotive industry, \cite{Jakubek:2002} uses PCA of training data to determine non-sparse areas of the measurement space. In the later phase, a distribution function in the PCA space is learned using neural networks.
Furthermore, \cite{ELGPN15} presents a PCA-based approach for condition monitoring of wind power plants. Following the PCA step, a distance of the new data to the normal data in the latent space is analyzed. 

\begin{figure}[tb]
  \begin{scriptsize}
    \centering  
    \def\svgwidth{200pt}
    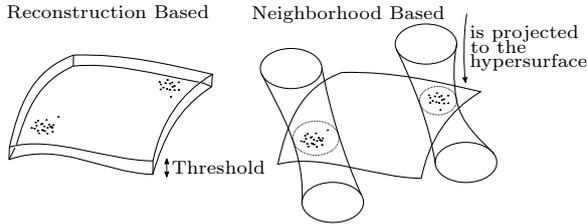
    \vglue 0ex plus 0.5ex minus 3.5ex
    {\caption{ Two aspects of modeling data using Autoencoder. The expected normal behavior volume in (\textit{Left}) the reconstruction based anomaly detection approach compared to a (\textit{Right}) neighborhood-based approach within the latent space.}\label{normalVolume}}
    \end{scriptsize}
  \end{figure}

However, linear techniques, such as PCA, often cannot adequately handle high dimensional complex data which exhibit nonlinear interrelations between variables. 
Following the similar motives as in the PCA case, we can develop autoencoder based anomaly detection \cite{Hawkins:2002}. 
However, as in the PCA case, if anomalies lay on the learned manifold, another, second phase approach must be applied in the transformed (latent) space. Clearly, to benefit from this, normal and anomalous data need to be separable in the lower dimensional embedding of the data (depicted in Figure \ref{normalVolume}).

%


\section{Proposed two-phase approach}
\label{sec:approach}
In a wind turbine system, power output increases at the cube of wind speed. Further, consider the time behavior of speed and power consumption of a conveyor. The two variables respond to the input command synchronously and, to some level, deterministically. Thus, their joint normal behavior is potentially one-dimensional issue ($x_1 = x_1(t)$, $x_2 = x_2(t)$). For both examples, any way of learning the system behavior from data should (implicitly) incorporate these nonlinear physical laws.

\begin{algorithm}
\caption{Two-phase anomaly detection concept}
\label{alg:ad}
\begin{algorithmic}[1]
\Statex \textbf{Input:} Learned AE/PCA model to reduce dimensionality ($f_{enc}$) and reconstruct ($f_{dec}$)
\Statex \textbf{Input:} Observation \textbf{x}
\Statex \textbf{Input:} Reconstruction error threshold $MSE_{th}$
\Statex \textbf{Outputs:} $Anomaly \in \{False,True\}$
\Statex
\State $\textbf{z}\gets f_{enc}(\mathbf{x})$
\State $\textbf{r} \gets f_{dec}(\mathbf{z})$

\State $E \gets MSE(\textbf{x}, \textbf{r}) $ according to Equation \ref{eq:l2loss}

\State $Anomaly1 \gets  E>MSE_{th}$

\State $Anomaly2 \gets $ Apply 2. phase approach given $\mathbf{z}$

\State $Anomaly \gets Anomaly1 \lor Anomaly2$

\\return $Anomaly$
\end{algorithmic}\label{alg:anomalydetection}
\end{algorithm}

We argue that dimensionality reduction using Autoencoder can capture the important aspects of the behavior of a CPPS, such as the aforementioned physics. On the other hand the encoded (latent) data representation can still be further analyzed. In this manner we propose a \textbf{two-phase anomaly detection concept} for CPPSs (Algorithm \ref{alg:ad}).

Once the Autoencoder model is learned from the data, anomaly detection is performed in the following way: In Steps $1-3$ reconstruction from the low-dimensional representation is compared to the input observation $\mathbf{x}$. The error is used as a measure of unconformity to the learned low-dimensional data manifold, which is hopefully an important aspect of the normal system behavior. Unconformity to the other aspects (see Figure \ref{normalVolume}) of the system behavior should be detected by a second-phase approach which operates in the autoencoder latent space (Step $5$). The overall decision is a disjunction of the decisions from two phases (Step $6$). The Autoencoder anomaly prediction is obtained by comparing the reconstruction error to a predefined threshold (Step $4$, see Figure \ref{anomalies}).
Below, we further analyze the concept and give a demonstration example.

\textit{\textbf{CPPS data characteristics.}} Typically, sensory data in CPPSs are given by noisy measurements of currents, power consumption, torque, positions, etc. Many such variables behave in certain deterministic ways and have definite physical relationships describing their joint behavior, some of which are linear 
while others are not
. Other variables include environmental influences that are less predictable. It is typical that such measurements have a limited number of modes in which the CPPS operates. A simple example would be a motor that is either running or turned off in which case the motor current would generate a two-modal distribution. Given several different signals, the overall distribution would show complicated clustering characteristics, the different modes extending in different dimensions wherever the signal is subject to continuous change or noise. 

Described characteristics are a motivation for a small demonstration example below. Consider, that we are focused on \textbf{behavior learning without knowledge about system structure} (causalities and other  variable dependencies).


%
  \begin{figure}[tb]
  \begin{scriptsize}
    \centering  
    \def\svgwidth{200pt}
    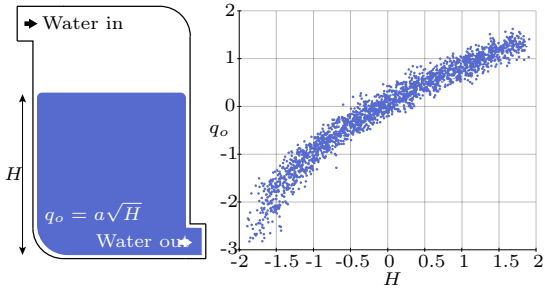
    \vglue 0ex plus 0.5ex minus 3.5ex
    {\caption{ \textit{Left:} Water tank system with two observed variables: water level $H$ and flow rate $q_o$ \textit{Right:} Plot of the observations of the two variables show nonlinear dependency between them.}\label{tank}}
    \end{scriptsize}
  \end{figure}

\begin{em}
\textbf{Water-tank system} is a simple system (see Figure \ref{tank} \textit{Left}) comprised of a tank filled with water. Close to the base of the tank is an opening through which the water can leave the tank. Two variables are observed in the system: water level ($H$) and flow rate out of the tank ($q_o$). The system behavior is simplified and the dependency between the two variables is described by the algebraic equation $q_o = a \sqrt H$ where $a$ is some constant. 

The underlaying normal behavior of the system is described by: 1) The water level is uniformly distributed: $H\sim\mathcal{U}(H_{min}, H_{max})$, the values out of this range are anomalies, 2) The flow rate is given by $q_o = a \sqrt H$.  3) Measurements of both variables add Gaussian noise to the nominal value (anomalies are out of the range of the Gaussian part). 
\end{em}

The system is depicted schematically in Figure \ref{tank}, including simulated data.
Observations of the water-tank system consist of two-dimensional real-valued vectors with components $H$ and $q_o$ at some time moment.



\textbf{\textit{What is a good representation of normal behavior to be learned?}}
When we manually model a system, the dynamics and interrelations between signals are given by physical relations, mechanical constraints, the solution of differential equations, etc. Each of these constraints reduces the intrinsic dimensionality of the data. In an ideal solution, those relations were learned and once some observed pattern does not satisfy the learned representation, a large anomaly score indicates the faulty event.

In a real physical system, signal observations are subject to noise, so a hard constraint softens, and can be defined as follows:
\begin{equation}
|f_\alpha(\mathbf{x})|\le\delta
\end{equation}
$f_\alpha$ is a scalar function defined on the space of observation vectors. For each constraint, there is one such equation, indexed by $\alpha$. The water-tank system constraint is given by $f(H,q_o)=q_o-a\sqrt{H}$. While $\delta=0$ constitutes the hard constraint, with $\delta>0$ we allow for some deviation from the physical law. All constraints taken together define the \textit{normal volume} of the data space.

\begin{assumption}[CPPS data intrinsic dimensionality]
We assume that $m$-dimensional observations from a CPPS have an intrinsic $p < m$.
\end{assumption}

\begin{figure}
\setlength{\unitlength}{1\columnwidth}
\begin{picture}(1.0,0.41)
\put(0.00,-0.03){\includegraphics[width=0.98\columnwidth]{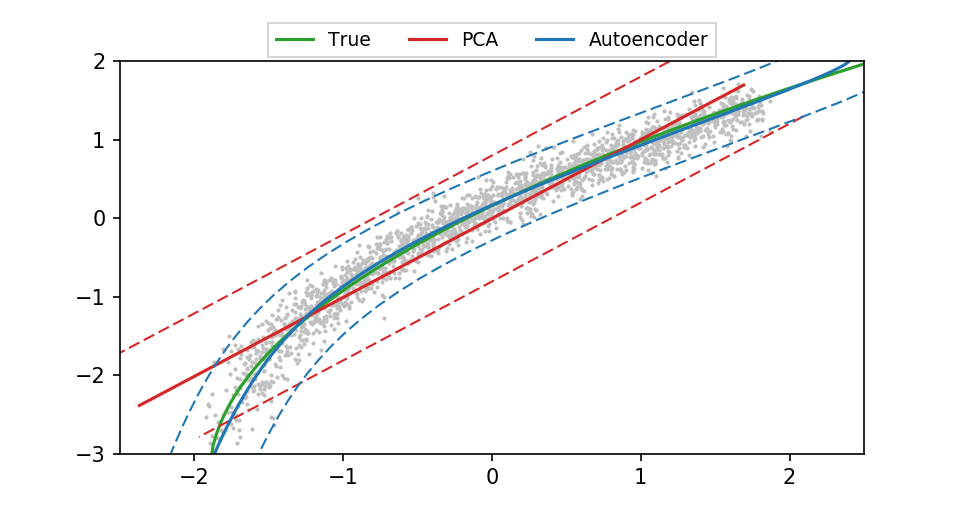}}
\put(0.01,0.23){\scriptsize{$q_o$}}
\put(0.49,-0.03){\scriptsize{$H$}}
\end{picture}
\caption{True underlying square root law (green), one dimensional representation of PCA (red) and autoencoder (blue). Dashed lines correspond to a region including 99.9\% of an independent test set (gray points). The yellow line represents a line perpendicular to the square root law.}\label{fig:watertank_dimred}
\end{figure}

Intrinsic dimensionality of the water-tank system is $p=\nobreak 1$, we therefore 
reduce the dimensionality to one dimension. The first principal component of the PCA and the learned Autoencoder representation are depicted in Figure \ref{fig:watertank_dimred}.


\textbf{\textit{Reconstruction error as the anomaly score.}} The reconstruction error is directly related to the soft constraint given by $\delta$. Geometrically, it can be interpreted as the squared distance of some point to the embedding of the learned manifold (with some caveats for non-optimal autoencoder). It is therefore a natural anomaly score which defines boundaries for the \textit{normal volume} of the data space in the directions that are not covered by the learned representation. 
In this example we fix the threshold at the 99.9\% quantile (thereby allowing a false positive rate of 0.1\%), see dotted lines in
Figure \ref{fig:watertank_dimred}.

It is obvious that a better representation (in case of the Autoencoder) will also yield a more fitting normal volume corresponding to a lower false negative rate. As expected, the Autoencoder captures the non-linear behavior and is able to learn the \textit{underlying physics} of the system in the domain where data is taken.


\textit{\textbf{Why a second phase approach?}} It should be noted that the learned manifold will generally span a larger section of data space than is actually populated. This is obviously true for the PCA where latent space is unconstrained. In the Autoencoder case this depends if any layer contains bounded activations such as hyperbolic tangent function. If not, latent space volume can also be infinite.
This also means that in general data will be sparsely distributed, because the true manifold is actually disconnected.
It is therefore often required to learn additional boundaries within the low dimensional manifold. (This situation is depicted on the right side of Figure \ref{normalVolume} for a $3\rightarrow 2$ reduction.)

This \textit{second phase} AD can consist of any established anomaly detection method which works well within a low dimensional space. A simple approach would be to set a threshold for each of the latent dimensions. This corresponds to a hypercube of normal volume in the latent space. 



In the following section we will investigate the discussed concepts and ideas with real-world data. This includes analysis of \textit{intrinsic dimensionality} as well as \textit{first} and \textit{second phase} anomaly detection (and the combination thereof) in realistic scenarios.
\section{Experiments}
\label{sec:evaluation}

\subsection{High Rack Storage System}\label{subsec:hrss}
\subsubsection{Data}
%
  \begin{figure}[tb]
  \begin{scriptsize}
    \centering  
    \def\svgwidth{230pt}
    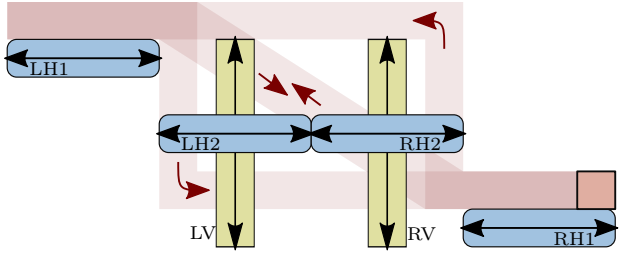
    \vglue 0ex plus 0.5ex minus 3.5ex
    {\caption{ High Rack Storage System with 4 horizontal and 2 vertical conveyors.}\label{hrss}}
    \end{scriptsize}
  \end{figure}

The High Rack Storage System (see Figure \ref{hrss}) is a demonstration system which transports objects between two different storage positions using four horizontal conveyors: LH1, LH2, RH1 and RH2 and two vertical conveyors: LV and RV which move LH2 and RH2, respectively. 

Each conveyor (drive) provides it's power consumption, voltage and speed.
Therefore, 18 signals in total are observed. A historical dataset contains 50502 observations of these signals during normal operation of the system: the object is transported from the bottom right to the top left position and back for 232 such cycles. When the object lays on two horizontal conveyors they are both running, otherwise only one conveyor is running. Two vertical conveyors are always running together trying to keep the same position. In different scenarios the object is moved following different paths as depicted in the figure.

HRSS dataset contains no anomalies so we can not evaluate the performance of anomaly detection approaches. However, HRSS is an interesting example from a CPPS domain which we can use to examine intrinsic dimensionality of the data.




\begin{figure}[tb]
\includegraphics[width=0.49\columnwidth]{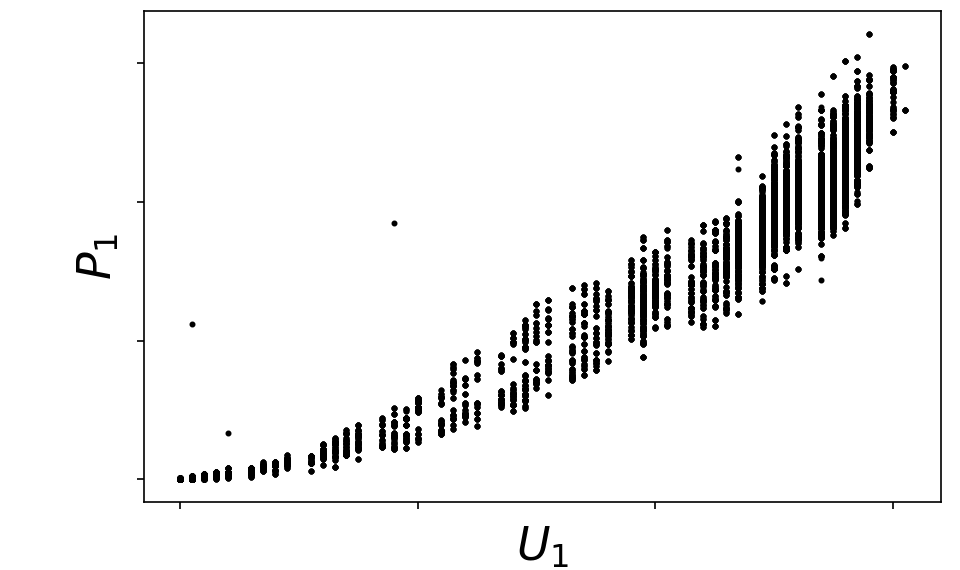}
\includegraphics[width=0.49\columnwidth]{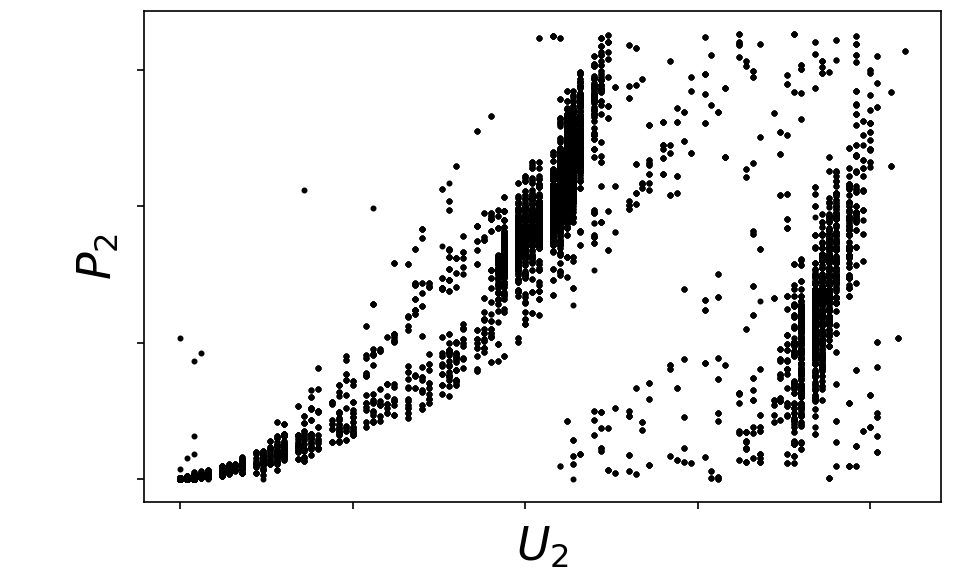}
\includegraphics[width=0.49\columnwidth]{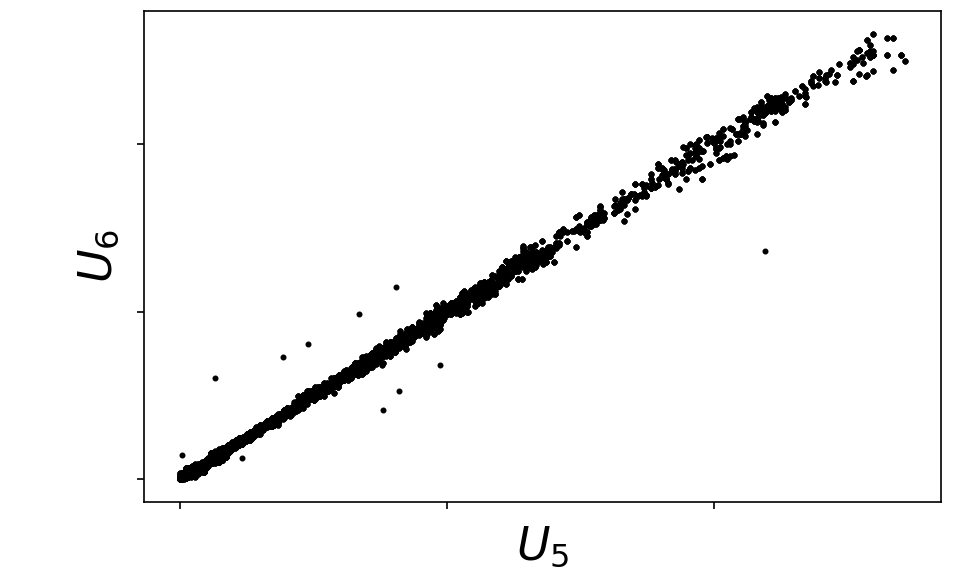}
\includegraphics[width=0.49\columnwidth]{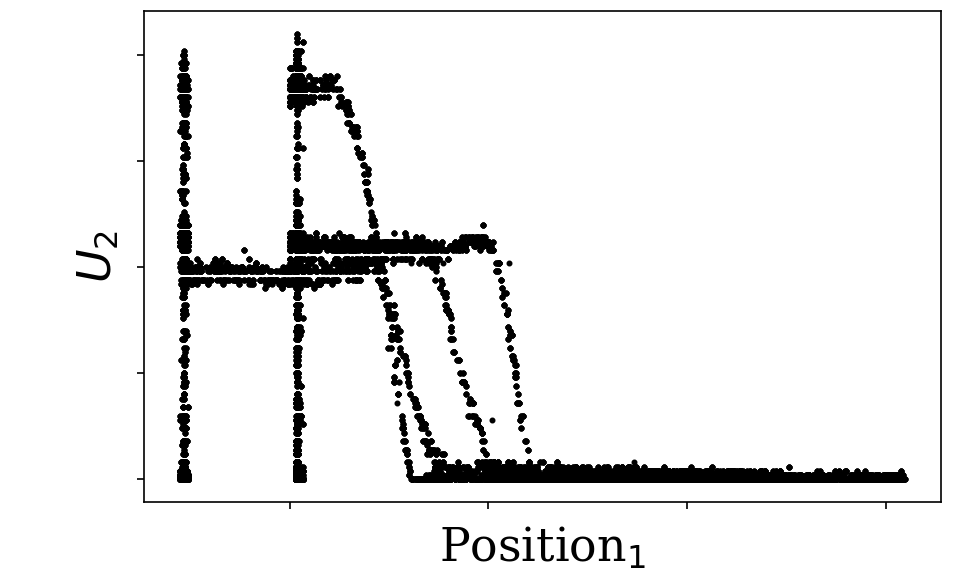}
\caption{\textit{Top-Left to Bottom-Right}: Voltage-Power plot of one of the six conveyor belts. Voltage-Power plot of an electrical drive carrying different loads back and forth. Voltage-Voltage plot of two drives moving in parallel. Voltage of one drive to the position of another drive.}\label{fig:HSS_scatter}
\end{figure}

\subsubsection{Discussion on the intrinsic data dimensionality}
Some typical patterns of correlation between different observables of one drive and between different drives are depicted in Figure \ref{fig:HSS_scatter}. The relationship between Voltage and Power is given by $P=U^2/R$. There exist no simple law between the position and the other independent variables, or different Voltages between the motors, but the physical constraints of the machine and its control system limit the possible space occupied by the normal behavior (see bottom of Figure \ref{fig:HSS_scatter}).
What can we tell about the intrinsic dimensionality of the data? There is a total of six not quite independent electrical drives, each providing observations about 3 interdependent variables. However, correlations between the drives' position and voltage/power is anything but simple.
A reasonable estimate of the intrinsic dimensionality would be the number of "independent" components, which is 6.

\subsubsection{Estimating the intrinsic data dimensionality using reconstruction error}
\begin{figure}
\centering
\includegraphics[width=0.80\columnwidth]{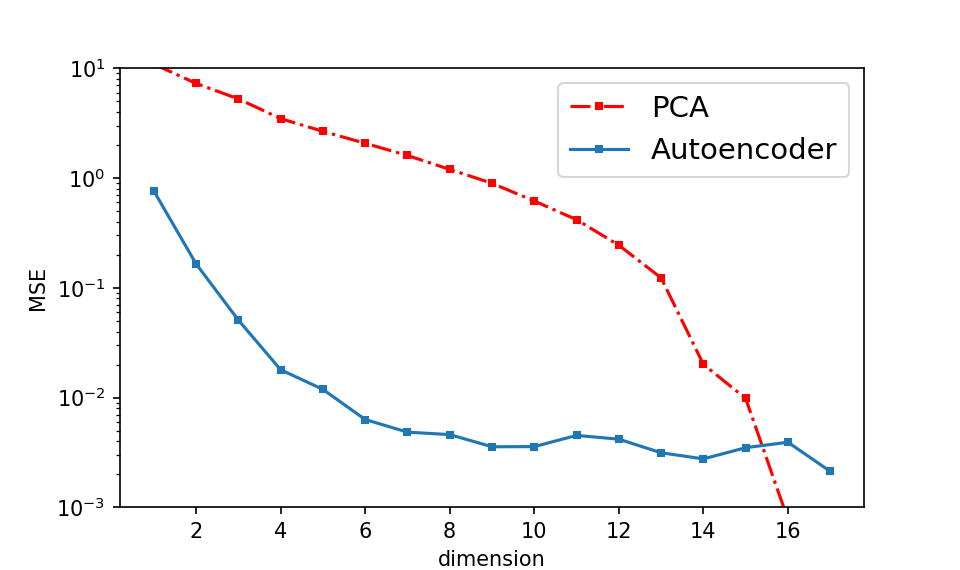}
\caption{Minimum reconstruction error on the independent test set of both PCA and Autoencoder by reducing the high rack storage system data from $18\rightarrow p$ dimensions.}\label{fig:HSS_recErr}
\end{figure}

To test the hypothesis, we train a PCA as well as several autoencoders with slightly different architectures and learning rates until reasonably well convergence has been achieved. The models are trained on the normalized data with zero mean and variance one (per signal) and tested on an independent test set with 5-fold cross-validation.
Good hyperparameter settings consist of three hidden layers for encoder/decoder with sizes between 30-200 and a learning rate of  $\sim 0.001$.
Figure \ref{fig:HSS_recErr} shows the total reconstruction error on the test set of both PCA and autoencoder model in a logarithmic scale. Values are taken to be the minimum (best model) from the cross-validation procedure.

For the Autoencoder, a significant drop below 1\% reconstruction error can be observed around reduced dimensionality $p=5\ldots8$ after which the curve flattens and increasing latent space dimensionality does not yield further improvement to the total reconstruction loss. This is not the case for the linear PCA, which only drops below 1\% (corresponding to a 99\% coverage of the variance) if 14 dimension are kept. This hints at a quite significant degree of non-linear relations between input features for the high rack storage data and lends credit to use non-linear approaches.

\subsection{MNIST}\label{subsec:mnist}
\subsubsection{Data}
Here we will compare the performance of first as well as second phase ($k$NN, $k$Means, One-Class-SVM) approaches with the MNIST handwritten data-set on an AD task. The data is high-dimensional, non-linear and can be said to have different modes (digits), which are typical CPPS-characteristics.
To perform anomaly detection, we construct 10 new datasets where each digit 0-9 is considered to play the role of the anomaly class. Training data will consist of all 50000 training samples minus the anomaly class. The test data (10000 samples) will remain as customary with labels one if the sample equals the anomaly class and zero otherwise.

\subsubsection{Evaluation Metrics}
In order to render the analysis independent of the chosen threshold we investigate the Area Under the Receiver Operator Characteristic (AU-ROC). A score of 0.5 indicates no better than random, while 1 signifies perfect performance.
The Autoencoder architecture remains constant throughout all MNIST experiments, a symmetric encoder-decoder setting with 3 hidden layers and intermediary representations of 256 neurons
[784, 256, 256, $p$, 256, 256, 784]. We use tanh activations except for both final layers of en- and decoder.

\begin{figure}
\includegraphics[width=0.98\columnwidth]{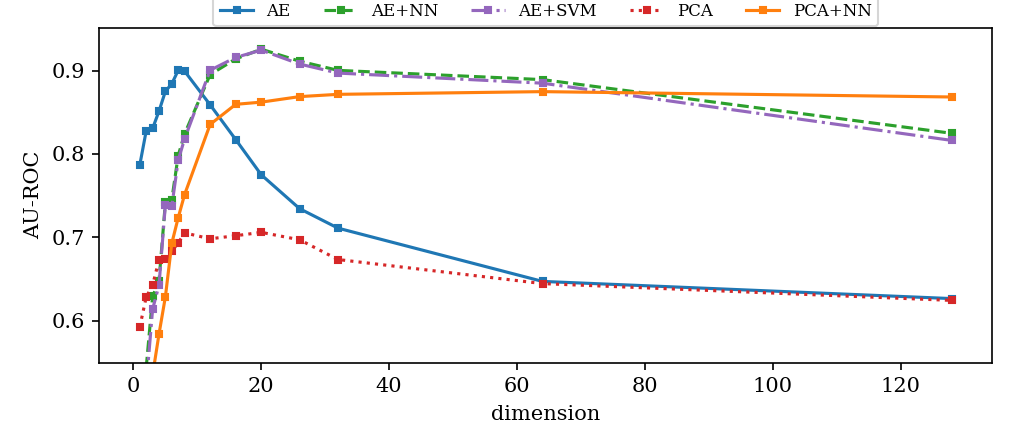}
\caption{Average AU-ROC scores over all anomaly classes (0-9).}\label{fig:au-roc_au-prc}
\end{figure}

\subsubsection{Analysis}
Figure \ref{fig:au-roc_au-prc} shows the AU-ROC against the size of reduced dimension.
First phase approaches work better for smaller dimensions compared to the second phase.
If the reduced space is too small, too much information is lost even to reconstruct the normal class. Here, both normal as well as anomalous classes are difficult to reconstruct, thereby lowering the discriminative power. If the reduced space is too large, both models use too large of a manifold to model the normal data. This leads to good reconstructions, even for previously unseen data originating from a different data distribution. Again, the discriminative power of the reconstruction error is reduced.

Second phases work best if the reduced dimensionality is not too small. Note that this number differs from the optimal dimensionality for the reconstruction-based AD approach which is evidence for a trade-off between the two phases when changing the size of the latent space.

\begin{table*}
\caption{Results of AU-ROC scores for all Approaches}
\label{tab:mnist}
\begin{center}
\begin{footnotesize}
\begin{sc}
\begin{tabular}{c | c c | c c}

With DR & Avg. AU-ROC & Dim. & W/O DR & Avg. AU-ROC\\ \hline\hline
AE & $0.90\pm 0.12$ & $7$ & & \\
PCA & $0.71\pm 0.16$ & $20$ & & \\
AE+NN & $\mathbf{0.93\pm 0.04}$ & $20$ & NN & $0.85\pm 0.16$\\
AE+$k$Mns & $0.61\pm 0.10$ & 6 & $k$Mns & $0.70\pm 0.16$\\
AE+SVM & $0.92\pm 0.05$ & $20$ & & \\

\end{tabular}
\end{sc}
\end{footnotesize}
\end{center}
\vskip -0.1in
\end{table*}

Table \ref{tab:mnist} summarizes the results of AU-ROC scores for all approaches, averaged over the anomaly classes (0-9) . The best dimensionality is chosen for each approach respectively. Second Phase experiments were done with $k$NN ($k=1$), One-Class-SVM ($\gamma=20/p$) and $k$Means ($k=9$) models. Anomaly scores are the distances to the nearest neighbor from the training set, the nearest cluster centroid and the support vector hyperplane, respectively.

The second phase approaches works better if the data has been reduced in dimensionality. By the reduction process, meaningful features for the task of discriminating against the unknown class have been extracted. However, this only applies if the DR technique was successful enough to capture the relevant feature dimensions that separate normal from anomalous data, which rarely happened with PCA  method. Here, DR plus second phase approach showed no significantly better than random chance at anomaly prediction.

\subsection{Wind power plant (WPP)}
We evaluate the proposed anomaly detection (AD) concept on a real-world wind power plant (WPP) use case presented in \cite{ELGPN15}. 
%
  \begin{figure}[tb]
  \begin{scriptsize}
    \centering  
    \def\svgwidth{220pt}
    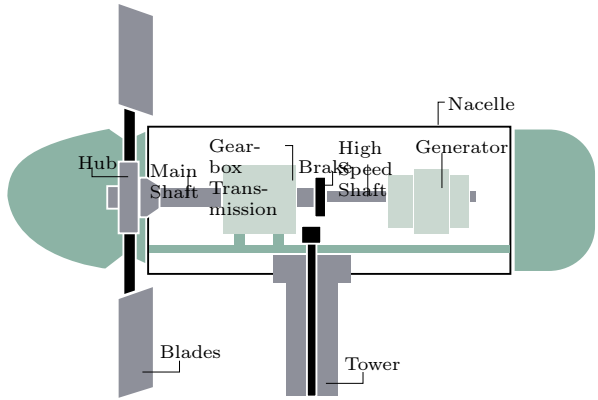
    \vglue 0ex plus 0.5ex minus 3.5ex
    {\caption{ Typical wind turbine parts.}\label{Windturbine}}
    \end{scriptsize}
  \end{figure}

Data are collected over a duration of 4 years, from a real-world WPP in Germany with 10 minutes resolution. The dataset consists of variables which describe the work environment (e.g. wind speed, air temperature) and the status of the plant (e.g. power capacity, rotation speed of generator, voltage of the transformer). Anomaly detection solution enables better maintenance, leading to a lower maintenance cost, reduced downtime and improved reliability and lifespan of the system.

A total of 12 variables are observed. The historical dataset is divided into a training set with 232749 observations and a test set with 11544 observations. The test set contains 4531 reported failures and 7013 observations considered normal.

Our experiments should answer three questions:

%
  \begin{figure}[b]
  \begin{scriptsize}
    \centering  
    \def\svgwidth{180pt}
    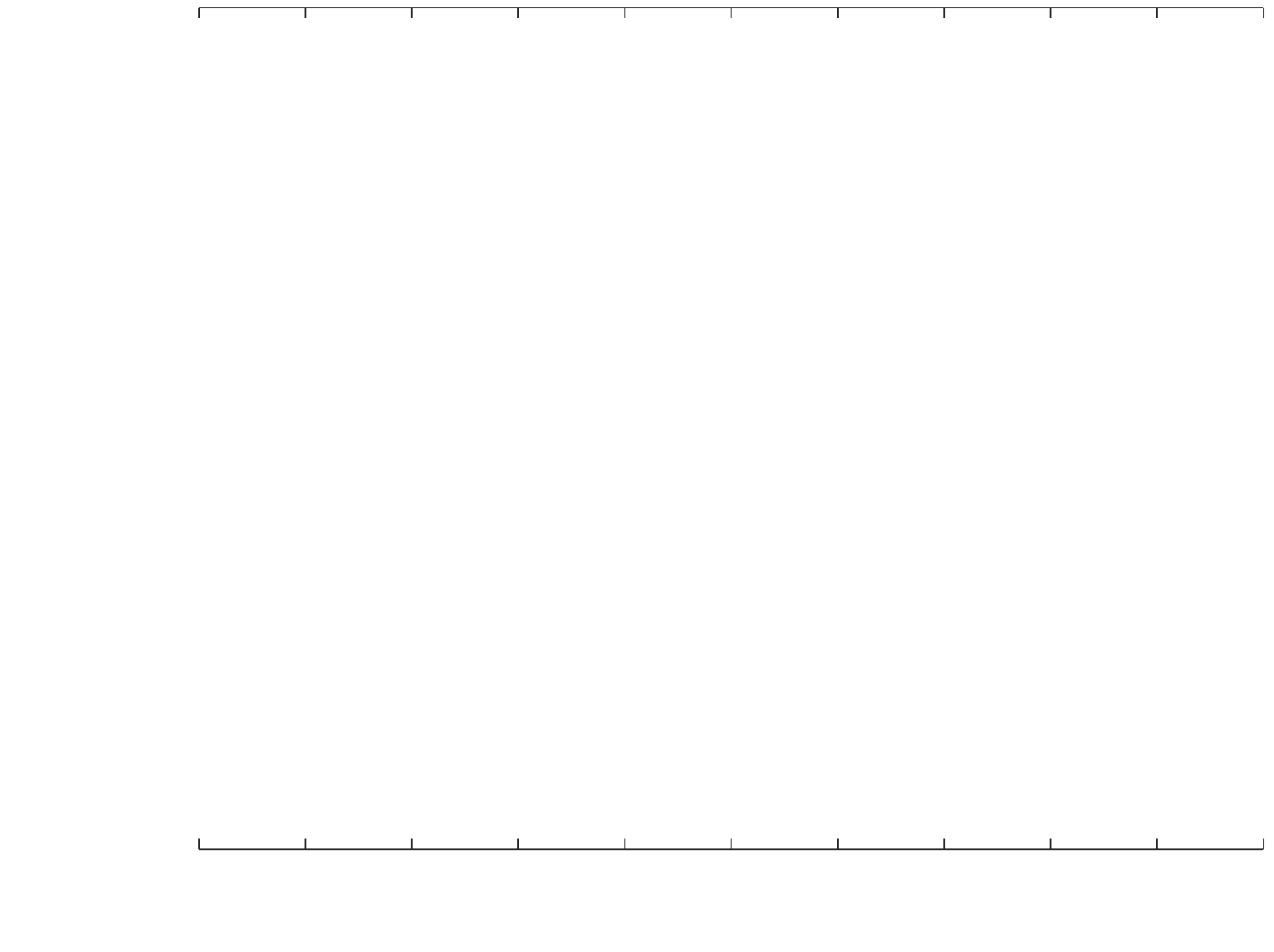
    \vglue 0ex plus 0.5ex minus 3.5ex
    {\caption{ Wind plant dimensionality reduction. MSE error for different $p$ (reduced dim.) for PCA and Autoencoder. We can estimate an intrinsic dimensionality 6 for this system.}\label{wind_MSE}}
    \end{scriptsize}
  \end{figure}

\noindent
\textbf{\emph{What can we say on the intrinsic data dimensionality?}}

\noindent
We experimented with different architectures (number of layers, layer sizes) for each $p = \{1,...,11\}$ (see Figure \ref{wind_MSE}). For each training of an autoencoder, optimization hyper-parameters were carefully selected. As there is an elbow at $p=6$ and the PCA performs significantly worse (except for $p\geq10$ where the learning of AE probably did not take long enough to reduce the error), we can claim nonlinear data manifold and intrinsic dimensionality $p=6$. 

\noindent
\textbf{\emph{What is the AD performance for different AE architectures when only DR is performed (no 2. phase)?}}

\begin{figure}[t]
  \begin{scriptsize}
    \centering  
    \def\svgwidth{180pt}
    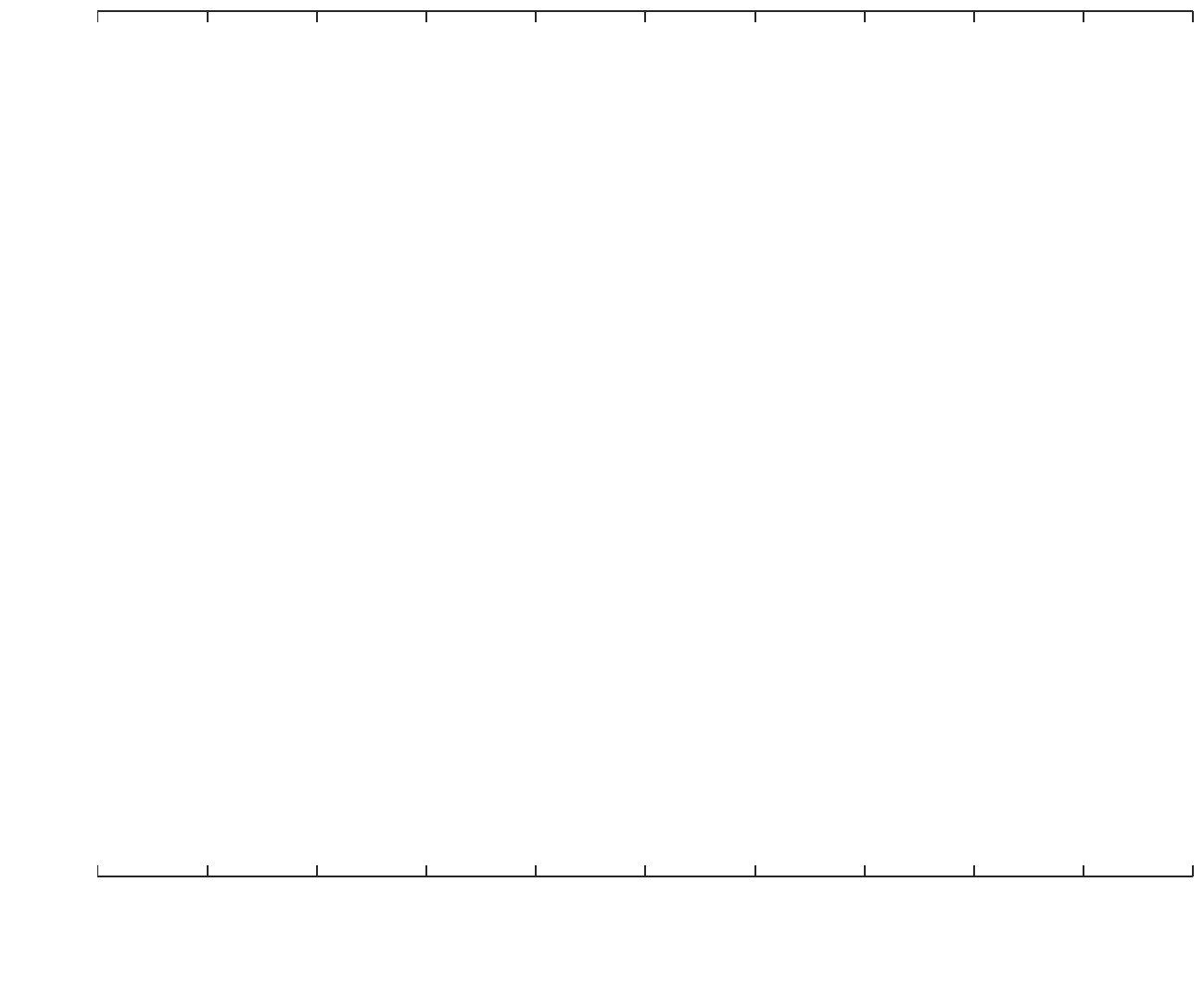
    \vglue 0ex plus 0.5ex minus 3.5ex
    {\caption{ AD F-score for different $p$ using only reconstruction error (blue line) and using 2-phase approach with kNN in the 2.phase (red line).}\label{wind_AD}}
    \end{scriptsize}
  \end{figure}

\noindent
Figure \ref{wind_AD} shows the performance of AE reconstruction-error-based approach for different $p$. The results for spectral clustering and PCA + kNN solution are taken from \cite{ELGPN15} (DBSCAN results were significantly worse and they are not presented on the figure). AE shows notable results of around $92\%$ when reduced dimensionality is equal or larger than the estimated dimensionality 6. This corresponds to the MSE for different $p$ (Figure \ref{wind_MSE}) which suggests expected correlation between MSE and anomaly detection performance for $p$ close to the intrinsic dimensionality.   

\noindent
\textbf{\emph{What changes if we include a 2. phase approach?}}

%

\noindent In Figure \ref{wind_AD} we see the performance of the 2-phase approach (see 
Algorithm \ref{alg:ad}) with kNN applied in the second phase. We can notice different effect of the autoencoder reduced dimensionality $p$ on the anomaly detection performance for AE only and AE+kNN cases. A two phase approach achieves best scores for the $p=9$. This larger $p$ was probably necessary for better separability of anomalies from the normal points in the latent space.

\section{Conclusion}
\label{sec:conclusion}


Anomaly detection (AD) has many applications in the domain of CPPS and beyond. Due to the curse of dimensionality, many prevailing algorithms cannot be used in the high dimensional space of input data.
By using dimensionality reduction, the optimization criterion -- usually MSE -- is itself an anomaly score that should be utilized for the classification process.

Furthermore, an independent \textit{second phase} AD approach can be used, that operates in the latent space.
We analyze this algorithm with respect to benefits for anomaly detection in CPPS scenario on several real-world data, using PCA and Autoencoder (linear and nonlinear) to perform the dimensionality reduction.
Results show that the second phase approach can benefit heavily from the DR-technique, outperforming its non-reduced baseline. This holds especially true for nonlinear DR.
This is because the curse of dimensionality could be partly overcome by learning meaningful features in a first step.

Furthermore, we observe a mismatch in the optimal dimension for independent \textit{first} and \textit{second} phase AD.
Second phase AD approaches perform better in a larger space compared to first phase approaches.
While increasing the latent space size, the general observation shows increasing AD-performance because more relevant features can be learned that are needed to discriminate the anomaly class. However, eventually this trend reverses because the latent space and therefore the expected \textit{normal volume} becomes so big that even anomalies are well represented.
In this case, anomalies are not sufficiently different in order to discriminate between them and the large variety of normal data. This seems to apply to both \textit{first}- and \textit{second stage} AD approaches.

It would be interesting to further study such phenomena on a larger scale with a variety of different CPPS data. This includes analysis of what kind of anomalies each phase can detect and how significant the overlap is, depending on latent space size. Finally, the prospect of a combination of first and second phase AD approaches into a joint end-to-end classification method should be investigated.

\bibliographystyle{IEEEtran}

\newcommand{\bslitpath}{Literature}

\bibliography{%
\bslitpath/diagnosis-lit,%
\bslitpath/learning-dm-lit,%
\bslitpath/lit,%
\bslitpath/misc-lit}

\end{document}